\documentclass{article}

\usepackage{iclr2027_conference,times}
\usepackage[T1]{fontenc}
\usepackage[utf8]{inputenc}
\usepackage{microtype}
\usepackage{amsmath,amssymb}
\usepackage{booktabs}
\usepackage{array}
\usepackage{graphicx}
\usepackage{subcaption}
\usepackage{xcolor}
\usepackage{enumitem}
\usepackage{placeins}
\usepackage{tikz}
\usetikzlibrary{arrows.meta,positioning,fit,calc}
\usepackage{hyperref}
\usepackage{url}
\usepackage[nameinlink,noabbrev]{cleveref}

\makeatletter
\renewcommand\section{\@startsection{section}{1}{\z@}%
  {-1.7ex plus -.4ex minus -.15ex}%
  {1.15ex plus .2ex minus .1ex}{\large\sc\raggedright}}
\renewcommand\subsection{\@startsection{subsection}{2}{\z@}%
  {-1.5ex plus -.4ex minus -.15ex}%
  {.6ex plus .15ex minus .1ex}{\normalsize\sc\raggedright}}
\renewcommand\paragraph{\@startsection{paragraph}{4}{\z@}%
  {.8ex plus .25ex minus .1ex}{-1em}{\normalsize\bfseries}}
\makeatother

\definecolor{m4lblue}{HTML}{2864B7}
\definecolor{m4lorange}{HTML}{D97732}
\definecolor{m4link}{HTML}{18212F}
\definecolor{m4lgrey}{HTML}{667085}
\definecolor{m4llight}{HTML}{EDF3FA}
\hypersetup{
  colorlinks=true,
  linkcolor=m4lblue,
  citecolor=m4lblue,
  urlcolor=m4lblue,
  pdftitle={How Far Should Tokenization Go? Predictive Effectiveness and Relational Losslessness},
  pdfauthor={Yi Wang}
}

\title{How Far Should Tokenization Go?\\
{\Large Predictive Effectiveness and Relational Losslessness}}
\iclrfinalcopy
\author{Yi Wang\\
Department of Electronic Engineering, Tsinghua University\\
\texttt{yiwang24@mails.tsinghua.edu.cn}\\[-1pt]
{\small Code: \href{https://github.com/kinssion/effectiveness-losslessness}
{\texttt{github.com/kinssion/effectiveness-losslessness}}}}
\date{}

\begin{document}
\maketitle
\fancyhead{}

\begin{abstract}
GPT-style models have achieved remarkable success with finite vocabularies of reusable tokens, making the token interface a central component of modern sequence modeling. Symbolic music appears naturally compatible with this paradigm: it consists of discrete note events and recurring structures such as chords, motifs, and phrases. However, when tokenization moves beyond language, the interface must be specified for each domain. Existing work offers many effective designs, but no unified criterion for deciding what tokenization should represent and how far it should go.

Using predictive codelength as a common criterion, we formulate the \emph{Effectiveness--Losslessness Framework} to define where tokenization should begin and where it should end. The \emph{Fact--Token Boundary} marks where observation-determined structure should enter the token interface, through operations such as coordinate construction. Within this interface, the resulting carrier may be reversibly recoded, for example through BPE, without changing the represented facts. The \emph{Token--State Boundary} marks where tokenization should stop: relations that depend on context should remain for model-state computation rather than being fixed in advance by the tokenizer.

We validate the framework through controlled multi-seed symbolic-music experiments, with an independent-corpus replication of the temporal intervention. Making musical time explicit consistently reduces predictive code and also improves pitch and duration prediction, while tonal-frame canonicalization and pitch factorization provide further gains. Fixed circle-of-fifths pitch coordinates instead increase predictive code, suggesting that imposing a fixed pitch relation before context can burden prediction. Reversible BPE substantially shortens the carrier but increases predictive codelength in every seed, showing that carrier compaction alone does not guarantee predictive gain.

Together, these results show that tokenization quality is determined neither by recoverability nor by sequence length, but by where predictive computation is placed. The framework therefore recasts tokenization as a model-relative boundary: static representation exposes reusable regularities, while contextual state resolves what remains context-dependent.
\end{abstract}

\section{Introduction}

\subsection{Tokenization in GPT-Style Models}

GPT-style models have achieved remarkable success by combining large-scale
sequence modeling with finite vocabularies of reusable tokens
\citep{vaswani2017attention,brown2020language}. The token interface maps text
into a discrete sequence of units that can be reused across contexts, making
tokenization a central component of the paradigm. Because language tokenizers
are mature and widely standardized, their design can appear secondary to the
model architecture.

\subsection{Symbolic Music and Reusable Structure}

Symbolic music appears naturally compatible with the same paradigm. It consists
of discrete Note and REST events and contains recurring structures such as
chords, motifs, phrases, and sections. Existing systems instantiate this idea
through event streams, compound words, learned segments, bar/position timing,
and absolute or relative pitch--time interfaces
\citep{huang2018musictransformer,huang2020pop,zeng2021musicbert,
fradet2023bpe,guo2023fme,inaba2024relativity}. Yet musical structure also spans
simultaneity, meter, transposition, and long-range recurrence, which are not
captured by a single serialization axis.

\subsection{The Missing Criterion for Tokenization}

Existing work offers many effective tokenization designs, but no unified
criterion for deciding what tokenization should represent and how far it should
go.

Comparisons between complete tokenizers cannot supply such a criterion.
Interfaces that preserve the same information may still impose different
predictive burdens on a finite model
\citep{xu2020usable,rajaraman2024analysis,fesharaki2026effective}.
Moreover, a change in predictive codelength may arise from coordinate
construction, reversible carrier coding, decoder or compute changes, or a
static relational prior. Recoverability and sequence length therefore do not
identify why one interface works better.

We address this problem by using predictive codelength as a common criterion
for measuring representation operations and determining where static
tokenization should begin and end.

\paragraph{Contributions.}
\begin{enumerate}[leftmargin=*,itemsep=2pt,topsep=3pt]

\item We make operation-level tokenization decisions measurable under a common
predictive-codelength criterion. By separating coordinate construction,
reversible carrier recoding, and fixed relational commitment, we define
operation-specific quantities that
identify how each intervention changes the predictive burden of a bounded
learner.

\item Building on this criterion, we formulate the
\emph{Effectiveness--Losslessness Framework} with two complementary principles.
Predictive Effectiveness defines the \emph{Fact--Token Boundary}, where
observation-determined structure should enter the token interface; Relational
Losslessness defines the \emph{Token--State Boundary}, where tokenization should
stop before context-dependent relations are fixed in advance.

\item We validate the resulting predictions through controlled symbolic-music
experiments. Observation-determined coordinate operations produce predictive
gains, fixed circle-of-fifths pitch coordinates increase predictive burden, and
reversible BPE substantially shortens the carrier without producing predictive
gain. An independent corpus replicates the temporal-coordinate direction.

\end{enumerate}

\section{The Effectiveness--Losslessness Framework}

\subsection{Predictive Codelength as a Common Criterion}

To decide how far tokenization should proceed, different representation
operations must be compared under a common predictive criterion. We use
predictive data codelength: an operation is evaluated by how it changes the
predictive burden of a specified learner under a fixed source and resource
budget.

We separate the declared fact surface, coordinate construction, carrier coding,
and contextual computation:
\begin{equation}
x_{\rm raw}
\xrightarrow{A}
x
\xrightarrow{R}
z_{1:N}
\xrightarrow{C}
u_{1:T}
\xrightarrow{F_\theta}
h_{1:T}.
\label{eq:pipeline}
\end{equation}
$A$ declares the modeled fact surface $x$, which may abstract the raw domain.
$R$ constructs coordinate carriers $z_{1:N}$, $C$ serializes or groups them as
$u_{1:T}$, and $F_\theta$ forms contextual states
$h_t=F_\theta(u_{<t})$ for $q_\theta(u_t\mid h_t)$.

For source distribution $P$, predictor family $\mathcal Q$, and resource budget
$\mathcal B$, we define
\begin{equation}
\mathcal L_{P,\mathcal Q,\mathcal B}(R,C)
=
\frac{
\inf_{q\in\mathcal Q_{\mathcal B}}
\mathbb E_{x\sim P}
\left[
-\sum_{t=1}^{T}\log_2 q(u_t\mid u_{<t})
\right]
+
\mathbb E_{x\sim P}[\ell_{\rm side}(x)]
}{
\mathbb E_{x\sim P}[N_{\rm fact}(x)]
},
\quad
u=C(R(x)).
\label{eq:codelength}
\end{equation}

The resource budget $\mathcal B$ fixes the learner and evaluation conditions
under which representations are compared, including model capacity, context,
training exposure, and selection policy. We measure predictive data codelength
rather than full MDL; model description length is outside this ledger.
Accordingly, all codelength differences are relative to the specified source,
predictor family, and budget \citep{xu2020usable}.

\paragraph{Matched intervention quantities.}

We distinguish three representation operations under the same predictive
criterion. Coordinate construction changes the model-facing coordinates,
carrier coding changes their reversible serialization, and relational
commitment imposes a fixed relation before contextual computation.

Using $\mathcal L$ as shorthand for fixed $(P,\mathcal Q,\mathcal B)$, we define
\begin{align}
G_{\rm coord}(R_0\!\rightarrow\!R_1)
&=
\mathcal L(R_0,C)-\mathcal L(R_1,C),
\nonumber\\
G_{\rm carrier}(C_0\!\rightarrow\!C_1)
&=
\mathcal L(R,C_0)-\mathcal L(R,C_1),
\nonumber\\
E_{\rm rel}(g)
&=
\mathcal L(R_g,C)-\mathcal L(R_{\rm open},C).
\label{eq:deltas}
\end{align}
Positive $G_{\rm coord}$ and $G_{\rm carrier}$ indicate predictive gains.
Positive $E_{\rm rel}$ indicates excess predictive code from the tested fixed
relation. Each quantity is interpreted only under a matched intervention in
which the corresponding operation changes and the remaining comparison
conditions are held fixed.

\subsection{Predictive Effectiveness: The Fact--Token Boundary}

The first decision is what structure should enter the token interface. The
\emph{Predictive Effectiveness Principle} admits an observation-determined
operation when exposing it reduces predictive codelength for the specified
learner.

Information preservation does not imply computational equivalence. Two
representations may preserve the same declared facts while imposing different
approximation burdens on a bounded learner. Invertibility is therefore
insufficient to establish coordinate usefulness.

The Fact--Token Boundary therefore marks where tokenization begins. A tested
candidate is predictively effective when it yields positive $G_{\rm coord}$
under a matched resource contract.

Two general coordinate operations are central in this work: decoupling and
denesting.

\paragraph{Decoupling.}
For a declared fact $x=(a,b)$, a recoverable factor-wise interface has
\begin{equation}
R_{\rm dec}(a,b)
=
\bigl(r_A(a),r_B(b)\bigr),
\qquad
\operatorname{Rec}_{\rm dec}\!\left(R_{\rm dec}(a,b)\right)
=
(a,b).
\label{eq:decoupling-map}
\end{equation}
The factors become separately addressable, but no independence is imposed:
$P(A,B)\neq P(A)P(B)$ may still hold. The operation removes only the
representational constraint that the two factors must be accessed through a
coupled coordinate, while leaving their statistical dependence available to the
contextual predictor. Its effectiveness requires matched positive gain
$G_{\rm coord}(R_{\rm coupled}\!\rightarrow\!R_{\rm dec})>0$; recoverability
alone does not determine the sign.

\paragraph{Denesting.}
For an observation-determined transformation $\phi$, a recoverable exposed
interface has
\begin{equation}
R_\phi(x)
=
\bigl(\phi(x),s_\phi(x)\bigr),
\qquad
\operatorname{Rec}_\phi\!\left(R_\phi(x)\right)
=
x.
\label{eq:denesting-map}
\end{equation}
Here $s_\phi(x)$ is the information required for inversion and may be empty
when $\phi$ is bijective; coded side information is charged through
$\ell_{\rm side}$ in \cref{eq:codelength}. Moving $\phi$ into the interface is
predictively effective only when
$G_{\rm coord}(R_0\!\rightarrow\!R_\phi)>0$ under the matched source, learner,
and budget. Determinism and reversibility alone do not imply this gain.

A particularly important case of denesting is \emph{canonicalization}, in which
$\phi$ aligns a known transformation orbit. This structured form removes
nuisance variation while preserving the information required to return to the
original coordinates, without assigning a contextual semantic relation.

These operations are candidates for the static interface because they are
determined by declared facts; their predictive value is measured, not assumed.

Deterministic relations may also enter the interface. Exact signed displacement
between preserved coordinates is observation-determined. The Fact--Token
Boundary is therefore not between facts and relations, but between structure
determined by observation and organization that requires context.

\subsection{Reversible Carrier Coding}

Once a coordinate interface is admitted, its contents may be serialized or
grouped in different reversible ways without changing the represented
coordinates. Common tokenization operations such as BPE act at this carrier
level by merging or regrouping symbols into a different token sequence.

Carrier compaction is not predictive compression:
\[
|C_1(z)| < |C_0(z)|
\;\not\Rightarrow\;
\mathcal L(R,C_1) < \mathcal L(R,C_0).
\]
A reversible recoding may shorten the emitted sequence while making prediction
no easier, or even harder, for the specified learner. Its value is therefore
measured by predictive carrier gain rather than sequence length alone.

\subsection{Relational Losslessness: The Token--State Boundary}

The second decision is where tokenization should stop. The
\emph{Relational Losslessness Principle} concerns organization whose useful
form cannot be determined from the preserved items alone.

For two items $x_i$ and $x_j$, suppose their useful relation depends on
surrounding context:
\begin{equation}
\rho_{ij}^{\mathcal C}
=
f(x_i,x_j,\mathcal C),
\qquad
\exists\,\mathcal C_1,\mathcal C_2:
\rho_{ij}^{\mathcal C_1}\neq\rho_{ij}^{\mathcal C_2}.
\label{eq:contextual-relation}
\end{equation}
The same preserved pair can therefore support different useful organizations
under different contexts.

A static interface makes a \emph{relational commitment} when it instead
privileges a context-free relation or topology
\begin{equation}
g_{ij}=g(x_i,x_j)
\label{eq:fixed-relation}
\end{equation}
before the relevant context is available. A \emph{relation-open} interface
preserves the identities or coordinates needed to infer such relations without
fixing this tokenizer-side geometry in advance.

Crucially, relational losslessness is not invertibility. A committed
representation may preserve every declared fact,
\[
R_g^{-1}(R_g(x))=x,
\]
yet still increase predictive burden:
\begin{equation}
R_g^{-1}(R_g(x))=x
\;\not\Rightarrow\;
E_{\rm rel}(g)\leq0.
\label{eq:invertible-not-relationally-lossless}
\end{equation}
Under the specified source, predictor, and budget, $E_{\rm rel}(g)>0$ is
\emph{relational coding loss}: the fixed commitment requires more predictive
code than its matched relation-open reference.

The \emph{Token--State Boundary} therefore marks where context-dependent
organization should remain unresolved by the static interface and instead be determined
through contextual state. This does not imply that every fixed relation
is harmful; a commitment is ruled out only when it produces relational excess
under the matched predictive criterion.

\subsection{The Unified Tokenization Boundary}

Together, the two principles define tokenization as a model-relative boundary
over predictive computation. Predictive Effectiveness moves
observation-determined structure into the static interface when doing so
reduces predictive burden, while Relational Losslessness stops this process
before context-dependent organization is prematurely fixed.

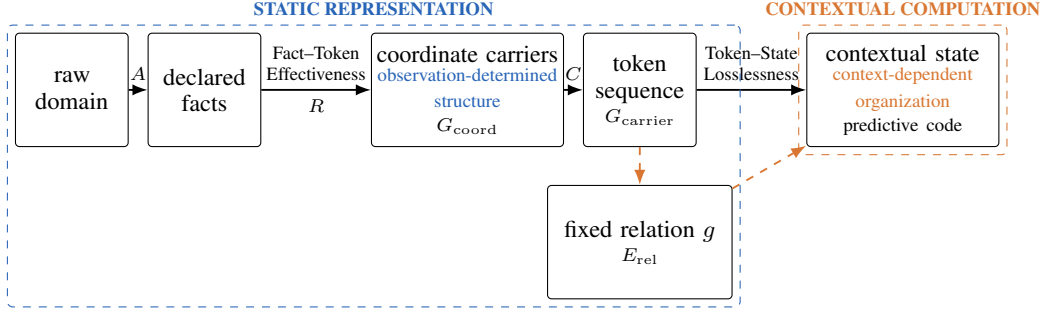
\begin{figure}[!htbp]
\centering
\begin{tikzpicture}[
    node distance=4mm and 2.5mm,
    font=\footnotesize,
    box/.style={draw,rounded corners=1.5pt,align=center,minimum height=15mm,
      text width=13.5mm,inner xsep=2pt},
    arrow/.style={-{Latex[length=2mm]},line width=0.7pt},
    side/.style={-{Latex[length=2mm]},line width=0.7pt,dashed,color=m4lorange}
]
\node[box] (raw) {raw domain};
\node[box,right=of raw] (facts) {declared facts};
\node[box,right=14.5mm of facts,text width=24mm] (coords) {coordinate carriers\\[-1pt]
  {\scriptsize\color{m4lblue}observation-determined\\structure}\\[-1pt]
  {\scriptsize $G_{\rm coord}$}};
\node[box,right=of coords] (tokens) {token\\sequence\\[-1pt]{\scriptsize $G_{\rm carrier}$}};
\node[box,right=14.5mm of tokens,text width=24mm] (state) {contextual state\\[-1pt]
  {\scriptsize\color{m4lorange}context-dependent\\organization}\\[-1pt]
  {\scriptsize predictive code}};
\node[box,below=5mm of tokens,text width=23mm] (rel) {fixed relation $g$\\[-1pt]{\scriptsize $E_{\rm rel}$}};
\draw[arrow] (raw) -- node[above,font=\scriptsize] {$A$} (facts);
\draw[arrow] (facts) -- node[above,font=\scriptsize,align=center]
  {Fact--Token\\Effectiveness} node[below,font=\scriptsize] {$R$} (coords);
\draw[arrow] (coords) -- node[above,font=\scriptsize] {$C$} (tokens);
\draw[arrow] (tokens) -- node[above,font=\scriptsize,align=center]
  {Token--State\\Losslessness} (state);
\draw[side] (tokens) -- (rel);
\draw[side] (rel) -- (state);
\node[draw=m4lblue,dashed,rounded corners=2pt,fit=(raw)(facts)(coords)(tokens)(rel),
  inner sep=1mm,label={[font=\bfseries\scriptsize,text=m4lblue]above:STATIC REPRESENTATION}] {};
\node[draw=m4lorange,dashed,rounded corners=2pt,fit=(state),inner sep=1mm,
  label={[font=\bfseries\scriptsize,text=m4lorange]above:CONTEXTUAL COMPUTATION}] {};
\end{tikzpicture}%
\caption{The Effectiveness--Losslessness Framework. Predictive Effectiveness
governs what enters the token interface, Relational Losslessness governs where
it stops, and reversible carrier coding is evaluated separately.}
\label{fig:v3-architecture}
\end{figure}

A token interface is therefore a carrier of admitted observable structure,
not a commitment to every higher-order organization that can be formed from
it. Reversible carrier coding acts within this interface and is evaluated
separately through $G_{\rm carrier}$. \Cref{fig:v3-architecture} summarizes the resulting division of predictive
computation across the static interface, carrier code, and contextual state.

\begin{quote}
\emph{Tokenization should expose reusable structure determined by observation
and stop before precomputing organization that context must determine.}
\end{quote}

\FloatBarrier

\section{Symbolic Music as a Controlled Testbed}

Symbolic music makes the framework concrete because score-level facts are
explicit while many useful organizations remain context-dependent. We use this
separation to derive coordinate, relational, and carrier predictions before
specifying their matched experimental realizations.

\subsection{Declared Fact Surface}

Under our score abstraction, each observed event is
\begin{equation}
    n_i=(t_i,o_i,p_i,d_i),
\end{equation}
where $t_i$ is event type, $o_i$ onset, $p_i$ pitch, and $d_i$ duration.
For a REST, onset and duration delimit the silent interval. These variables
define the declared fact surface.

Expressive performance and low-level MIDI attributes lie outside this
abstraction. All modeled non-drum notes form one unlabeled event stream.
Same-onset events remain distinct, and serialization order assigns neither
melodic priority nor voice identity.

\subsection{Predictive Effectiveness in Symbolic Music}

\paragraph{Musical time.}
Sequence position reflects serialization rather than musical time. We instead
expose each onset through a directed multiscale metric coordinate
\begin{equation}
\tau_i=\phi(o_i)
=
\bigl(
\text{bar progress},
\text{beat/bar/multibar phases}
\bigr),
\label{eq:musical-time}
\end{equation}
and derive exact signed temporal displacements between events from their
observed onsets. These operations expose score-determined temporal structure
without assigning any contextual musical organization.

Predictive Effectiveness predicts positive coordinate gain when this exposure
makes reusable temporal regularities easier for the bounded predictor to
access. Because time participates in the joint event distribution, the gain
need not be confined to Time prediction; it may also appear in pitch, duration,
or event type.

\paragraph{Pitch factorization.}
Absolute pitch can be written bijectively as pitch class and register,
\begin{equation}
    p_i\longleftrightarrow(c_i,r_i),\qquad
    c_i=p_i\bmod 12,\qquad
    r_i=\left\lfloor p_i/12\right\rfloor.
\end{equation}
This decouples two reusable factors without assuming that they are statistically
independent. Predictive Effectiveness therefore asks whether making them
separately addressable lowers predictive codelength; invertibility alone does
not determine the answer.

\paragraph{Tonal canonicalization.}
Transposition defines a known transformation orbit over absolute pitch.
Choosing a deterministic reference frame gives
\begin{equation}
    \widetilde p_i=p_i-s,\qquad
    p_i=\widetilde p_i+s,
\end{equation}
with the inverse shift retained outside the predictor. This is a structured
denesting operation: it removes a nuisance degree of freedom without supplying
tonic, mode, scale degree, or harmonic function as semantic labels.

The predicted signature is positive coordinate gain after charging the
information required for inversion.

\subsection{Relational Losslessness in Symbolic Music}

Pitch provides a controlled test of the Token--State Boundary because pitch
identity can be preserved while its model-facing geometry is changed. A
relation-open interface exposes pitch class $c_i$ as a categorical identity
without prescribing a fixed distance between pitch classes.

A relation-committed interface instead maps pitch class to a fixed circular
coordinate
\begin{equation}
\gamma_\kappa(c_i)
=
\left[
\sin\frac{2\pi\kappa(c_i)}{12},
\cos\frac{2\pi\kappa(c_i)}{12}
\right],
\label{eq:fixed-pitch-topology}
\end{equation}
where $\kappa(c)=c$ gives chromatic order and
$\kappa(c)=7c\bmod 12$ gives circle-of-fifths order. This mapping remains
injective, but its geometry privileges proximity under the chosen topology
before musical context is available.

Relational Losslessness predicts $E_{\rm rel}>0$ when such a fixed geometry
precommits pitch organization that would be better resolved from context.

\subsection{Reversible Carrier Coding in Symbolic Music}

The same admitted musical coordinates can be serialized through different
reversible carriers. Event symbols such as bar, position, pitch, and duration
may be emitted directly or regrouped by a learned merge scheme such as BPE
without changing the recoverable coordinate grammar.

The carrier hypothesis is deliberately weaker than a compression claim:
shortening the emitted sequence need not reduce predictive codelength. The
appropriate empirical signature is therefore $G_{\rm carrier}$, evaluated with
the represented coordinates held fixed.

\section{Related Work}

\begingroup
\setlength{\parskip}{0pt}

\paragraph{Predictive tokenization under bounded models.}

Tokenization and representation mappings have been formalized independently of
any particular vocabulary \citep{gastaldi2025foundations}. Predictive usable
information and bounded-model analyses show that information-preserving
representations can impose different predictive burdens on computationally
bounded learners
\citep{xu2020usable,rajaraman2024analysis,fesharaki2026effective,
bechler2026lost}. Related work also shows that token count or sequence
compaction alone does not determine predictive quality
\citep{zouhar2023noiseless,schmidt2024morethancompression}.
These results motivate predictive codelength as our accounting criterion; our
focus is the operation-level distinction between coordinate construction,
carrier recoding, and relational commitment.

\paragraph{Musical coordinates and relational structure.}

Symbolic-music models expose bar/position timing, temporal and pitch relativity,
and combinations of absolute and relative musical attributes
\citep{huang2018musictransformer,huang2020pop,fradet2023timeduration,
guo2023fme,inaba2024relativity,guo2025moonbeam}. These works provide precedents
for placing domain structure into the model-facing interface. We distinguish
observation-determined coordinate exposure from context-free relational
commitment and evaluate the two through matched predictive interventions.

\paragraph{Carrier coding and sequence compaction.}

Compound words, BPE, and learned merging change how musical information is
carried by the token sequence
\citep{zeng2021musicbert,fradet2023bpe,qu2024mupt,
geerlings2020gpt2,pasquier2025midigpt}. Such recoding can alter sequence length,
vocabulary, and the effective resources available per declared fact. We treat
reversible carrier coding as a separate operation: represented coordinates are
held fixed, and shorter serialization is not assumed to imply lower predictive
codelength.

\paragraph{Compression and contextual abstraction.}

Hierarchical music models organize computation across contextual scales, while
segment-based tokenizers move higher-level abstractions into the token interface
\citep{yu2022museformer,huang2025musetok}. DeCo similarly separates token
compression from semantic abstraction in multimodal modeling
\citep{yao2024deco}. These works motivate the broader allocation question
addressed here: which structure should be made static and reusable, and which
organization should remain for contextual state.

\endgroup

\section{Experiments and Discussion}
\label{sec:v3-experiments}

\subsection{Matched Experimental Design}

We instantiate the three operation classes in \cref{eq:deltas} through matched
Pop1K7 comparisons:

\begin{center}
\begin{tabular}{lll}
\toprule
Quantity & Comparison & Deliberate change \\
\midrule
$G_{\rm coord}$ & A--D & musical-time coordinates \\
$G_{\rm coord}$ & D--H & pitch factorization \\
$G_{\rm coord}$ & F--G & tonal canonicalization \\
$E_{\rm rel}$ & D--E & fixed chromatic geometry \\
$E_{\rm rel}$ & H--I & fixed fifths geometry \\
$G_{\rm carrier}$ & J--K & reversible BPE recoding \\
\bottomrule
\end{tabular}
\end{center}

Within A--D, A uses generic sequence position, B exposes absolute musical
position, C replaces it with multiscale musical time, and D additionally exposes
exact temporal displacement. The remaining comparisons change only the
operation named above within their matched family.

Only matched within-family deltas receive operation-level interpretation;
cross-family scores are descriptive. Full representation, resource, parameter,
and compute contracts are given in \cref{app:protocol,app:tokenizer-controls}.

Pop1K7 \citep{huang2020pop} is split by source-song lineage before windowing.
All primary comparisons use three preregistered seeds, validation-only
checkpoint selection, and one sealed clean-test pass. Codelength is normalized
by declared Note/REST facts and includes EOS and required side information;
G is charged four inverse-shift bits per independent window. Full accounting
and integrity contracts appear in \cref{app:protocol}.

\subsection{Predictive Coordinate Gain}

\begin{figure*}[t]
\centering
\includegraphics[width=0.92\textwidth]{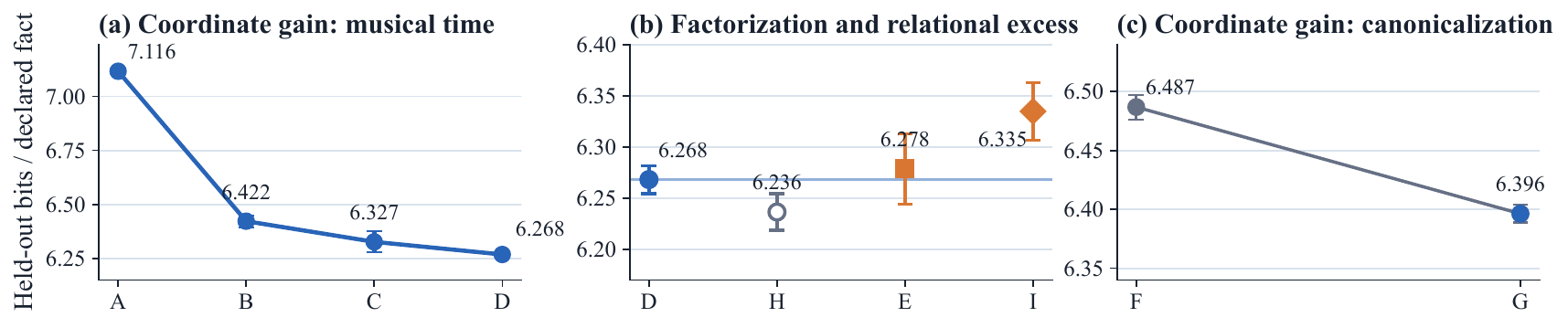}
\caption{Pop1K7 clean-test codelength per declared musical fact. Error bars show
sample SD over three preregistered seeds. Panel (a) reports the staged A--D
musical-time family. Panel (b) contains separate matched comparisons: D/H
estimates pitch-factorization gain; D/E and H/I estimate relational excess for
chromatic and fifths commitments, respectively. Panel (c) reports the matched
F/G canonicalization family. G
includes a conservative four-bit side code per window.}
\label{fig:v3-representation-operations}
\end{figure*}

\paragraph{Musical time.}
Codelength decreases at every stage of A--D, with gains of $0.694$, $0.096$,
and $0.059$ bits/fact for A$\rightarrow$B, B$\rightarrow$C, and
C$\rightarrow$D. Overall,
\[
G_{\rm coord}(A\!\rightarrow\!D)=0.84809\ \text{bits/fact},
\]
an $11.92\%$ reduction, with all three seeds favoring D
(\cref{fig:v3-representation-operations}a).

Only $0.287$ bits/fact of this gain comes from Time prediction; roughly two
thirds of the gain occurs in the other event factors, principally Pitch and
Duration. Thus
exposing musical time reorganizes the joint conditional distribution rather
than merely making timestamps easier to predict. The complete decomposition
appears in \cref{app:factor-decomposition}.

The A/D direction also replicates in all three ComMU seeds
\citep{lee2022commu}, with a mean gain of $0.111$ bits/fact. Because the runs
were still improving at the budget limit, this establishes directional rather
than converged replication (\cref{app:commu-replication}).

\paragraph{Pitch-coordinate operations.}
After charging the inverse shift, tonal canonicalization yields
$G_{\rm coord}(F\!\rightarrow\!G)=0.09055$ bits/fact, with all three seeds
improving. The operation removes a transposition degree of freedom without
supplying tonic, mode, scale degree, or harmonic function. Pitch factorization
yields a smaller gain of $0.03171$ bits/fact; two seeds improve and one is
approximately tied (\cref{fig:v3-representation-operations}b,c).

\subsection{Relational Excess Codelength}

The primary Token--State test compares relation-open H with the fixed
circle-of-fifths geometry I. Although both preserve pitch class and register,
the fixed topology incurs
\[
E_{\rm rel}(\text{fifths})
=\mathcal L(I)-\mathcal L(H)
=0.09831\ \text{bits/fact},
\]
with all three seeds regressing. Pitch contributes $0.08428$ bits/fact of this
excess, locating most of the penalty in the representation that was changed.

The chromatic D/E comparison is much smaller
($E_{\rm rel}=0.01032$ bits/fact) and has mixed seed directions. It therefore
shows no reliable gain from the tested fixed chromatic geometry, rather than
the seed-consistent loss observed for fifths.

A separate fixed-target probe provides complementary state-side evidence:
removing either ordered context increases code by about $0.27$ bits/event,
while shuffling the same content increases it by $0.669$ bits/event
(\cref{app:state-context}). This supports context-sensitive use of preserved
evidence but is not a matched Token--State intervention.

\subsection{Reversible Carrier Coding}

The J/K comparison holds the recoverable coordinate grammar fixed while K
applies reversible BPE to J's serialized stream. BPE reduces emitted targets
from $3.563$ to $1.538$ per fact ($56.8\%$), yet every seed requires more
predictive code:
\[
G_{\rm carrier}(J\!\rightarrow\!K)
=-0.81061\ \text{bits/fact}.
\]
Thus substantial carrier compaction does not imply predictive compression under
the tested learner and budget. Full carrier-induced vocabulary, parameter, and
compute changes are reported in \cref{app:tokenizer-controls}.

\subsection{Scope and Limitations}

All effects are predictive-data-codelength results relative to the tested
source, learner, and resource budget, and mechanistic attribution is restricted
to matched families. The results do not imply that all fixed relations or BPE
schemes are harmful; ComMU provides directional rather than converged
replication, and the context probe is supporting evidence only. We do not
establish perceptual quality, interpretable hidden-state variables, or
asymptotic scaling.

\section{Conclusion}

How far should tokenization go? Our central answer is that tokenization is a
model-relative allocation of predictive computation. The
\emph{Effectiveness--Losslessness Framework} makes this allocation operational:
Predictive Effectiveness admits observation-determined structure when exposing
it reduces predictive codelength, while Relational Losslessness stops static
representation before context-dependent organization is fixed in advance.
Reversible carrier coding is evaluated separately. Tokenization quality is
therefore determined neither by recoverability nor by sequence length alone,
but by where predictive computation is placed.

Symbolic music provides controlled evidence for each part of this distinction.
Musical-time coordinates reduce predictive code by $11.92\%$, with roughly two
thirds of the gain appearing outside Time prediction; tonal canonicalization
remains beneficial after its inverse is charged, and pitch factorization yields
a smaller gain. In contrast, an injective fixed circle-of-fifths geometry
produces seed-consistent relational excess, while reversible BPE shortens the
carrier by $56.8\%$ yet increases predictive codelength by $0.811$ bits/fact.
An independent corpus reproduces the temporal-coordinate direction. Together,
these results support an operation-level view of tokenization: expose reusable
structure, preserve contextual relational freedom, and do not equate carrier
compaction with predictive compression.

These results suggest a revealing view of representation: tokens compress what
can be standardized in advance, while contextual states learn how to compress
what must remain context-dependent. From this perspective, self-attention is
powerful not merely because it accesses context, but because it can construct
the predictive representation itself as context changes.

\section*{AI Use Statement}

Generative AI systems, including OpenAI Codex/ChatGPT and Google Gemini,
were used as research-assistance tools for conceptual brainstorming and
critical discussion, code drafting and refactoring, experiment orchestration,
literature discovery, figure and table preparation, and editorial revision.
AI outputs were not treated as scientific evidence. The author retained
final control over the research questions, experimental protocols, data
inclusion, interpretations, and claims; verified the cited sources,
implementations, and reported results; and take full responsibility for the
submission. Formal results were produced through the frozen deterministic
protocols and auditable receipts reported in the paper rather than from
unverified model-generated values.

\section*{Reproducibility Statement}

Reproducibility is addressed in \cref{sec:v3-experiments} and
\cref{app:protocol,app:tokenizer-controls,app:state-context,app:commu-replication}.
They specify data lineage, declared facts, numerator and denominator accounting,
side-information codes, matched interventions, active parameters, budgets,
seeds, and checkpoint rules. Protocol, source-tree, split-manifest, and
checkpoint hashes bind each result to its frozen implementation and data
lineage. Test data were unavailable to training and checkpoint selection;
the sealed test was materialized once after checkpoint lock. The J/K technical
continuation used only hash-verified caches and did not reopen the test manifest
or row file.

\section*{Ethics and Data Statement}

The primary representation and carrier-coding experiments use Pop1K7
\citep{huang2020pop}; ComMU \citep{lee2022commu} provides an independent
temporal-coordinate replication. Both are used for non-commercial research with
attribution, and we make no claim of ownership over source compositions.
No human-subject study or listener data collection was conducted. Public
data-derived artifacts will follow the applicable upstream terms.

\bibliographystyle{iclr2027_conference}
\bibliography{references}

\clearpage
\appendix

\FloatBarrier
\section{Frozen Protocol Summary}
\label{app:protocol}

\begin{table}
\centering
\caption{Frozen source, learner, resource, and accounting contract.}
\begin{tabular}{@{}p{0.23\linewidth}p{0.70\linewidth}@{}}
\toprule
Item & Frozen value \\
\midrule
Dataset
& Pop1K7 constant-4/4 piano arrangements; source-song split before windowing \\

Songs / windows
& 1,378 / 18,460 train; 184 / 2,011 validation; 185 / 1,979 sealed test \\

Carrier and target
& A--I: exact Note/REST stream with a separate termination target and
Type$\rightarrow$Time$\rightarrow$Pitch$\rightarrow$Duration factors; J/K:
lossless \textsc{Bar}/\textsc{Position}/pitch-or-\textsc{Rest}/duration stream,
with train-only reversible BPE in K \\

Accounting
& numerator: all Note/REST factor bits, EOS overhead, and required per-sample
side bits; denominator: declared Note/REST facts only \\

Backbone
& 16 Transformer layers, width 64, 4 heads, FFN 128, dropout 0; J/K use the
same causal backbone with a serialized-token input/output head \\

Parameters
& 815,571 total; effective trainable counts are A 809,475; B 811,079;
C 810,499; D/F/G 812,815; E/I 805,455; H 813,519. Vocabulary-dependent J/K
totals are 689,349 / 804,675 \\

Context
& deterministic complete 8-bar windows, capped at 2,048 note tokens; no random
note or bar crop \\

Seeds
& 20260819, 20260820, 20260821 \\

Budget
& 2,000 target-equivalent epochs; 3,650,482,000 full-split exposures;
2,664,430,000 matched-subset exposures. Coordinate/relation families fix
decoder and compute policy per fact; J/K fix original-event exposure and the
backbone while carrier length and vocabulary change as the intervention \\

Optimization
& AdamW; $3\times10^{-4}$ to $3\times10^{-5}$ cosine schedule with warmup;
batch size 128 songs \\

Selection
& minimum native micro validation code; ties favor the earliest exposure.
Fact-level rebasing and G's constant side charge do not change checkpoint order \\

Side information
& G shift support $\{-5,\ldots,6\}$: fixed-length 4-bit code, charged once per
8-bar window in the primary ledger and once per source song in the amortized audit \\

Admissible deltas
& coordinate: A--D, D--H, F--G; relation: E--D, I--H; carrier: J--K only.
J--D and K--D are descriptive full-interface comparisons \\

Test firewall
& test data disabled and unloaded during training and checkpoint selection;
one manifest and row-file read after checkpoint lock; J/K cache-only technical
continuation without a second test access \\
\bottomrule
\end{tabular}
\end{table}

The A--E, F--G, H--I, and J--K families each have frozen protocol, source-tree, and
manifest hashes in the repository receipts. $P$ is the corresponding frozen
split, $\mathcal Q$ is the shared causal-Transformer family, and $\mathcal B$
is the tabled resource contract. Inactive representation paths remain
instantiated but frozen; active parameter count is therefore part of the full
interface intervention.

\FloatBarrier
\section{Exposure-Matched Optimization Dynamics}
\label{app:learning-dynamics}

\begin{figure*}[!htbp]
\centering
\includegraphics[width=0.98\textwidth]{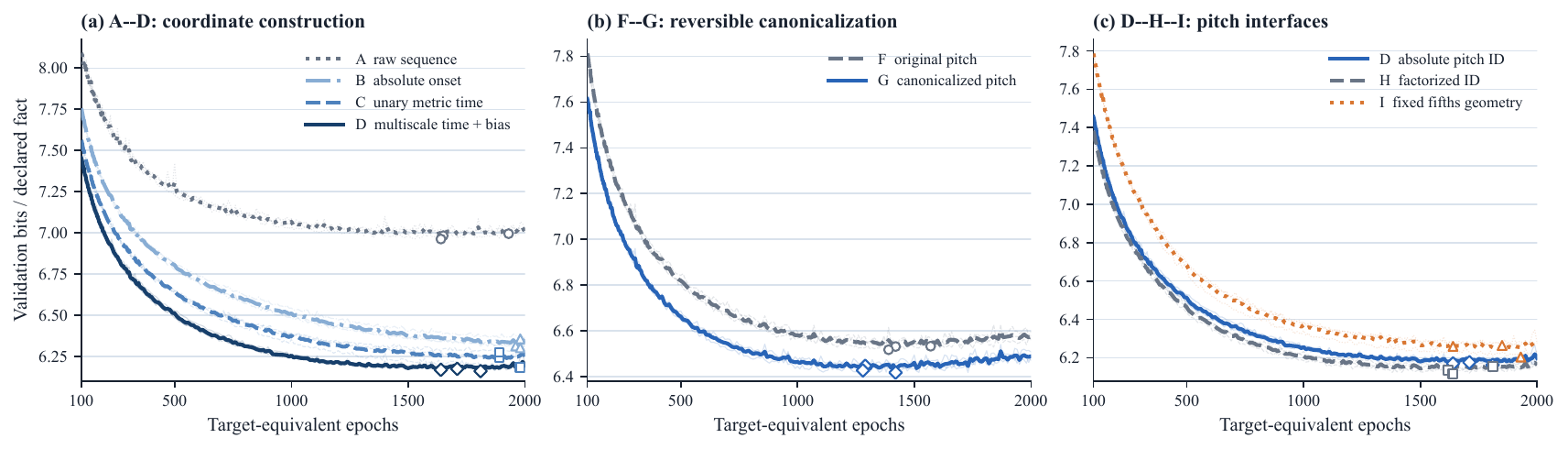}
\caption{
Exposure-matched validation dynamics for the three frozen Pop1K7
representation families. Thin lines show individual preregistered seeds;
thick lines show the three-seed mean; open markers identify the independently
validation-selected checkpoint for each seed. The plot begins at 100
target-equivalent epochs to remove the initialization transient, while all
source observations remain in the released curve table. Every run continues
to the same 2,000-epoch maximum budget. A--D and D--H--I use the full split;
F--G use their exactly matched high-confidence subset. Curves are rebased to
declared facts; G includes the constant four-bit side charge per window. No test
value enters training, plotting, or checkpoint selection.
}
\label{fig:app-learning-dynamics}
\end{figure*}

The figure is included to distinguish achievable predictive code length from
transient optimization effects. Main results therefore use the minimum
validation code length under the common maximum budget, followed by one sealed
test evaluation of the frozen checkpoint; they do not compare arms at an
arbitrary common epoch.

\FloatBarrier
\section{Full A--I Arm Table}

\begin{table}
\centering
\scriptsize
\caption{
Frozen Pop1K7 representation arms. A--E and H--I use the full original-pitch
split; F--G use the exactly matched high-confidence tonal-frame subset.
Values are bits per declared fact; G includes the conservative per-window side
code. Absolute scores are compared only within matched families.
}
\begin{tabular}{@{}p{0.15\linewidth}cp{0.35\linewidth}rr@{}}
\toprule
Family & Arm & Active interface & Validation & Held-out test \\
\midrule
Coordinate: time
& A
& raw pitch identity; generic sequence position
& $6.98064\pm0.01633$
& $7.11618\pm0.00765$ \\

Coordinate: time
& B
& A + learned absolute onset/bar lookup
& $6.32519\pm0.02835$
& $6.42223\pm0.02730$ \\

Coordinate: time
& C
& raw pitch; unary ordered and multi-scale metric time replacing
generic/absolute position features
& $6.23396\pm0.04671$
& $6.32662\pm0.04850$ \\

Coordinate: time
& D
& C + pairwise onset, meter, bar, and beat relations
& $6.16952\pm0.00681$
& $6.26809\pm0.01361$ \\

Relation: chromatic
& E
& D time; chromatic-circle phase + absolute register replacing categorical
pitch-class identity
& $6.18494\pm0.03436$
& $6.27841\pm0.03446$ \\
\midrule

Coordinate: canonicalization
& F
& original absolute pitch on the matched music21 high-confidence subset
& $6.52727\pm0.00848$
& $6.48686\pm0.01038$ \\

Coordinate: canonicalization
& G
& deterministic whole-song tonal canonicalization; inverse retained outside model
& $6.43069\pm0.01506$
& $6.39631\pm0.00744$ \\
\midrule

Coordinate: pitch factorization
& H
& learned categorical absolute pitch class + learned absolute register; D time
& $6.13428\pm0.01959$
& $6.23638\pm0.01785$ \\

Relation: fifths
& I
& fixed circle-of-fifths phase + learned absolute register; D time
& $6.24231\pm0.03362$
& $6.33469\pm0.02811$ \\
\bottomrule
\end{tabular}
\end{table}

F and G use the same songs and non-pitch fields from the music21 consensus
high-confidence subset. F presents original absolute pitch. G estimates one
deterministic whole-song reference frame and canonicalizes pitch according to
\begin{equation}
    \widetilde{p}_i=p_i-s,
\end{equation}
predicts $\widetilde{p}_i$, and restores physical pitch through
\begin{equation}
    p_i=\widetilde{p}_i+s.
\end{equation}
The reference shift is retained for deterministic reconstruction but is not
supplied to the Transformer. Dataset key labels, tonic, mode, and harmonic
function are unused.

The manifest contains 12 possible shifts, so a fixed-length code costs four
bits. The primary result charges this once for each of 1,500 test windows; a
source-song decoder can instead reuse one shift across the 132 test songs.

\begin{table}
\centering
\small
\caption{Clean-test F/G side-information ledger in bits per declared fact.}
\begin{tabular}{@{}lrrr@{}}
\toprule
Ledger & F code & G code & $G_{\rm coord}(F\!\rightarrow\!G)$ \\
\midrule
Model code only & 6.48686 & 6.36174 & 0.12512 \\
4 bits / 8-bar window & 6.48686 & 6.39631 & 0.09055 \\
4 bits / source song & 6.48686 & 6.36478 & 0.12208 \\
\bottomrule
\end{tabular}
\end{table}

\FloatBarrier
\section{Full Paired Factor Decomposition}
\label{app:factor-decomposition}

\begin{table}
\centering
\scriptsize
\caption{
Held-out paired changes in bits per declared fact. Negative values favor the
left arm. Class is assigned by intervention; Side is nonzero only for G.
}
\begin{tabular}{@{}llrrrrrr@{}}
\toprule
Comparison & Class & Total & Type & Time & Pitch & Duration & Side \\
\midrule
B$-$A & $G_{\rm coord}$ & $-0.693948$ & $-0.020283$ & $-0.248170$
      & $-0.243551$ & $-0.181944$ & $0$ \\

C$-$B & $G_{\rm coord}$ & $-0.095607$ & $-0.000855$ & $-0.033450$
      & $-0.052890$ & $-0.008412$ & $0$ \\

D$-$C & $G_{\rm coord}$ & $-0.058531$ & $-0.000165$ & $-0.005304$
      & $-0.043773$ & $-0.009290$ & $0$ \\

D$-$A & $G_{\rm coord}$ & $-0.848087$ & $-0.021304$ & $-0.286924$
      & $-0.340213$ & $-0.199646$ & $0$ \\

E$-$D & $E_{\rm rel}$ & $+0.010321$ & $-0.000059$ & $-0.006559$
      & $+0.017385$ & $-0.000445$ & $0$ \\

G$-$F & $G_{\rm coord}$ & $-0.090548$ & $+0.000003$ & $+0.002641$
      & $-0.137238$ & $+0.009474$ & $+0.034573$ \\

H$-$D & $G_{\rm coord}$ & $-0.031711$ & $+0.000028$ & $-0.004057$
      & $-0.025560$ & $-0.002121$ & $0$ \\

I$-$H & $E_{\rm rel}$ & $+0.098307$ & $-0.000013$ & $+0.010218$
      & $+0.084278$ & $+0.003824$ & $0$ \\

I$-$D & descriptive & $+0.066596$ & $+0.000015$ & $+0.006161$
      & $+0.058718$ & $+0.001703$ & $0$ \\
\bottomrule
\end{tabular}
\end{table}

\FloatBarrier
\section{Split and Test Integrity}

Pop1K7 is split by source song before windowing. Exact,
transposition-invariant, and rhythm/interval duplicate components remain within
one split; all audited cross-split counts are zero. Model selection uses
validation only, with test disabled during training.

All 27 frozen checkpoints were evaluated once with no parameter updates. The
full test split and matched F/G subset were opened only for this pass.

\FloatBarrier
\section{Representation Details}

All modeled non-drum notes are placed in a single unsegregated event stream.
No melody, bass, harmony, accompaniment, instrument-part, or voice partition
is supplied to the model. Notes sharing an onset remain separate observable
events and are serialized by ascending MIDI pitch. There is no cardinality
head, pitch-set decoder, onset-group identity, melody extraction, chord label,
or voice label.

For events without pitch, including REST events, a fixed type-specific
tie-breaking convention is used. Serialization order is deterministic but does
not imply performance order, melodic priority, voice membership, or semantic
importance.

Duration remains an exactly recoverable attribute of each Note or REST event
rather than an active-note state reconstructed from later transitions. The
decoder order is
\begin{equation}
    \textsc{Type}
    \rightarrow
    \textsc{Time}
    \rightarrow
    \textsc{Pitch}
    \rightarrow
    \textsc{Duration}.
\end{equation}
This is a chain-rule factorization of the exact-event distribution rather than
a semantic ranking of musical attributes.

C, D, E, H, and I use directed bar progress together with within-beat,
beat-within-bar, bar, four-bar, and sixteen-bar unary phases. The directed
component disambiguates repeated periodic phases, so the joint musical-time
coordinate uniquely determines the supported score onset:
\begin{equation}
    \tau_i=\phi(o_i),
    \qquad
    \phi^{-1}(\tau_i)=o_i.
\end{equation}
The unary interface is therefore an invertible coordinate transformation that
preserves the supported score onset.

D, E, H, and I additionally use signed onset displacement, within-bar
displacement, signed bar distance, and signed beat distance as deterministic
attention biases. These pairwise quantities are computed from preserved
musical-time coordinates and do not assign chord, motif, phrase, or voice
identity.

D uses raw MIDI-pitch identity. H uses categorical pitch class
\begin{equation}
    c_i=p_i\bmod 12
\end{equation}
together with categorical absolute register
\begin{equation}
    r_i=\left\lfloor\frac{p_i}{12}\right\rfloor.
\end{equation}
The pair $(c_i,r_i)$ bijectively determines the supported absolute pitch.
H is relation-open in the experimental sense: pitch-class identities have
learned categorical embeddings, and no context-free pitch-class distance kernel
is inserted.

E replaces categorical pitch-class identity with the chromatic phase
\begin{equation}
    \gamma_{\mathrm{chr}}(p_i)
    =
    \left[
        \sin\!\left(\frac{2\pi(p_i\bmod12)}{12}\right),
        \cos\!\left(\frac{2\pi(p_i\bmod12)}{12}\right)
    \right]
\end{equation}
and retains absolute register.

I uses the same phase construction after mapping pitch class to the
circle-of-fifths index
\begin{equation}
    f_i
    =
    \bigl(7(p_i\bmod12)\bigr)\bmod12,
\end{equation}
giving
\begin{equation}
    \gamma_{\mathrm{fif}}(p_i)
    =
    \left[
        \sin\!\left(\frac{2\pi f_i}{12}\right),
        \cos\!\left(\frac{2\pi f_i}{12}\right)
    \right].
\end{equation}
In E and I, phase inner products privilege chromatic or fifths proximity before
contextual attention. The phases remain injective on the twelve supported pitch
classes; their intervention class follows from the fixed topology, not from the
sign of the result.

F and G use identical songs and non-pitch fields from the music21
high-confidence subset. F predicts original absolute pitch. G applies
tokenizer-side whole-song canonicalization,
\begin{equation}
    \widetilde{p}_i=p_i-s,
\end{equation}
predicts canonical pitch, and applies evaluator-side reconstruction,
\begin{equation}
    p_i=\widetilde{p}_i+s.
\end{equation}
The reference shift is retained for deterministic reconstruction but is not
supplied to the Transformer. Dataset key labels, tonic, mode, scale degree, and
tonal function do not enter G. The primary ledger assigns $s$ a four-bit code
for each independent window; the source-song audit amortizes the same code over
all windows from one song.

\FloatBarrier
\section{State-Side Use of Preserved Relations}
\label{app:state-context}

\begin{figure}
\centering
\includegraphics[width=0.86\linewidth]{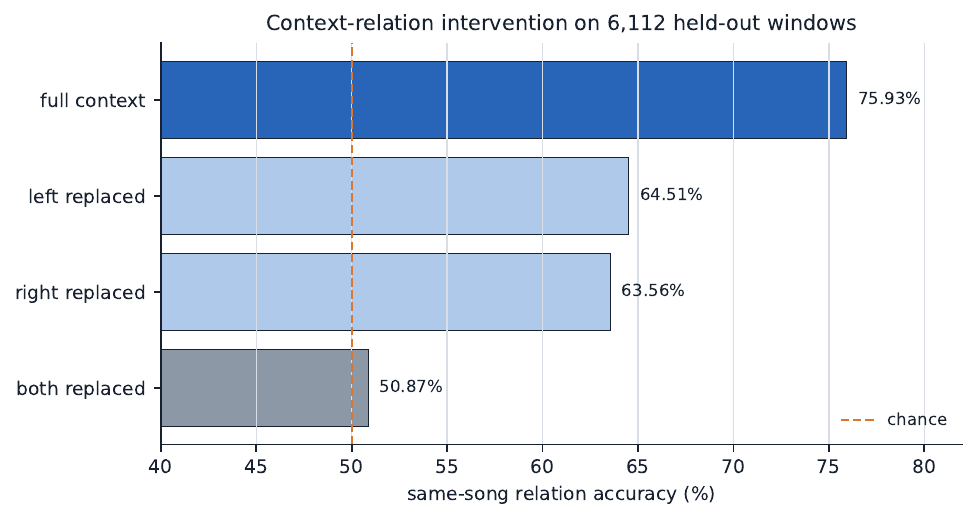}
\caption{
Secondary context-relation probe. Replacing either side degrades compatibility
prediction; replacing both approaches the target-only control.
}
\label{fig:context}
\end{figure}

An earlier relation probe, separate from A--I, tests whether a candidate
four-bar target belongs between left and right context. On 6,112 windows from
99 held-out POP909 songs, full-context accuracy was $75.93\%$, compared with
$48.87\%$ for the target-only condition. Replacing left or right context
reduced accuracy by 11.42 and 12.37 percentage points, respectively, while
replacing both reduced accuracy to $50.87\%$
(\cref{fig:context}).

Because this probe uses an earlier composition-level representation, it is
supporting evidence for contextual dependence rather than part of the primary
tokenization result. It is not pooled numerically with the main fixed-target
intervention and is not used to support any representation ranking.

A separate fixed-target intervention preserves the target and varies only its
available context. Its native evaluation units are retained because this probe
uses a different source and task; the values are not pooled with A--I.

\begin{table}
\centering
\small
\caption{Fixed-target context intervention in its native predictive units.}
\label{tab:app-context-intervention}
\begin{tabular}{@{}lrr@{}}
\toprule
Condition & bits/event & change from Full \\
\midrule
\textsc{Full} & $3.5394$ & --- \\
\textsc{Left} & $3.8083$ & $+0.2689$ \\
\textsc{Right} & $3.8117$ & $+0.2723$ \\
\textsc{Shuffled} & $4.2084$ & $+0.6690$ \\
\bottomrule
\end{tabular}
\end{table}

Removing either ordered side costs about $0.27$ bits/event; shuffling the same
amount of content costs $0.6690$. The state therefore uses ordered
content--time binding rather than context quantity alone.

\FloatBarrier
\section{Evidence Boundaries}

\begin{itemize}
    \item
    The reported data term is a resource-bounded predictive code length, not
    full MDL or an unrestricted coding optimum. Every comparison is relative to
    its declared model, optimizer, context, budget, and fact denominator.

    \item
    Fact abstraction, reversible carrier coding, and state-side relational
    commitment are different interventions. We assign a result by what changes,
    and compare only matched arms within the corresponding family.

    \item
    J/K is the admissible carrier intervention: source split, seeds, backbone,
    original-event exposure, and validation-only selection are fixed while
    reversible coding changes carrier length and vocabulary. J--D and K--D are
    descriptive only because event/field autoregression, decoder factorization,
    Transformer positions per fact, and computation per fact are unmatched.

    \item
    H improves over D in two seeds and ties in one. E is mixed; I regresses in
    all three seeds. These results concern the tested factorization and fixed
    topologies under slightly different active parameter counts, not every
    possible pitch geometry.

    \item
    F/G uses a matched high-confidence subset and a deterministic whole-song
    frame. The primary ledger charges four side-information bits per independent
    window; F/G is not ranked against full-split D and does not test online frame
    inference.

    \item
    Temporal gains establish predictive utility for the tested deterministic
    time coordinates and relations. The context probes establish sensitivity to
    ordered content--time binding. Neither identifies a named phrase, cadence,
    motif, chord, or voice variable in the hidden state.

    \item
    Exact recovery is claimed only for the declared Note/REST fields and, where
    applicable, their separately coded inverse. It does not imply byte-level
    recovery of raw MIDI, controller data, instrumentation, micro-timing, or
    acoustic performance.
\end{itemize}

\FloatBarrier
\section{Pop1K7 Fixed-Coordinate Carrier-Coding Control}
\label{app:tokenizer-controls}

The frozen serialized evaluator appends one EOS target to every window and
includes that target in each arm's loss. Its archived denominator of $244{,}248$
likewise includes $1{,}979$ termination entries. The paper ledger preserves each
arm's complete numerator, including its own EOS cost, and rebases only the
denominator to the $242{,}269$ declared Note/REST facts:
\begin{equation}
\mathcal L^{\rm paper}
=\mathcal L^{\rm archive}\frac{244{,}248}{242{,}269}.
\end{equation}
No equality of EOS overhead across J, K, or D is assumed.

\begin{table}[!htbp]
\centering
\small
\caption{Pop1K7 fixed-coordinate carrier-coding control. Serialized-target
ratios count one EOS target per window; predictive bits include EOS and are
normalized by declared Note/REST facts. Values are mean $\pm$ sample SD over
three preregistered seeds.}
\label{tab:app-tokenizer-controls}
\begin{tabular}{@{}lrr@{}}
\toprule
Interface & serialized targets/fact & held-out bits/fact \\
\midrule
J: uncompressed serialized coordinate stream & 3.563 & $5.96305\pm0.10235$ \\
K: J + train-only reversible BPE & 1.538 & $6.77367\pm0.05643$ \\
$K-J$ & $-2.025$ & $+0.81061\pm0.15007$ \\
\bottomrule
\end{tabular}
\end{table}

J is an event-preserving REMI-like stream of \textsc{Bar}, exact
\textsc{Position}, pitch/\textsc{Rest}, and duration symbols; it is not a
no-coordinate baseline. K applies train-only reversible BPE without changing
the recoverable fields or coordinate grammar. The split, seeds, 16-layer
backbone, original-event exposure, and validation-only checkpoint rule are
matched. Carrier-dependent vocabulary heads bring total parameters to 689,349
for J and 804,675 for K and are part of the tested recoding intervention.

The paired $K-J$ penalties are $0.95195$, $0.82678$, and $0.65312$ bits/fact.
Thus
$G_{\mathrm{carrier}}(J\!\to\!K)=\mathcal L(J)-\mathcal L(K)
=-0.81061$ bits/fact: a $56.8\%$ target reduction produces negative carrier
gain under this source, learner family, and budget. Validation selection used no
test data; the clean-test manifest and row file were each read once. A technical
continuation evaluated hash-verified caches only, with no checkpoint update or
test-time BPE fitting.

\paragraph{Descriptive cross-interface reference.}
J is below D in every seed, whereas K is above D in every seed. These complete
system scores are reported transparently but are not admissible coordinate or
carrier deltas.

\begin{table}[!htbp]
\centering
\small
\caption{Descriptive Pop1K7 full-interface comparisons in the common paper
ledger. Differences are left minus right; values are mean $\pm$ sample SD.}
\label{tab:app-jk-descriptive-reference}
\begin{tabular}{@{}lrrl@{}}
\toprule
Comparison & mean difference & seed direction & interpretation \\
\midrule
J--D & $-0.30504\pm0.11508$ & 3/3 J below D & descriptive only \\
K--D & $+0.50558\pm0.04366$ & 3/3 K above D & descriptive only \\
\bottomrule
\end{tabular}
\end{table}

J and D differ in event- versus field-level autoregression, decoder
factorization, Transformer positions and contextual recomputations per original
fact, and whether coordinates enter as serialized content or a side interface.
J is therefore a strong serialized coordinate interface, not evidence for any
single one of those mechanisms. Full-system codelength ranks the tested systems;
matched interventions determine why they differ.

\FloatBarrier
\section{Independent-Corpus A/D Replication}
\label{app:commu-replication}

The independent replication uses 9,299 ComMU songs in 4/4, split into 8,652
train, 323 validation, and 324 sealed-test songs. Scores retain their original
C-major/A-minor normalization; no second transposition is applied. A and D use
the same normalized pitch representation and differ only in their
musical-time interface. They otherwise share targets, model envelope, decoder,
optimizer, batch construction, context policy, three preregistered seeds, five
target-equivalent epochs, and the validation-only checkpoint rule within this
corpus. The ComMU budget is not asserted to equal Pop1K7 exposure.

\begin{table}[!htbp]
\centering
\small
\caption{Frozen ComMU 4/4 A/D replication. Predictive bits include sequence
termination and are normalized by declared Note/REST facts. Summary rows report
mean $\pm$ sample SD over three preregistered seeds; lower is better.}
\label{tab:app-commu-replication}
\begin{tabular}{@{}lrrr@{}}
\toprule
Arm/seed & validation & clean test & test D$-$A \\
\midrule
A: raw sequence & $12.56786\pm0.14342$ & $12.33993\pm0.14311$ & --- \\
D: musical-time coordinates & $\mathbf{12.39563\pm0.15434}$ & $\mathbf{12.22855\pm0.18073}$ & $-0.11139$ \\
\midrule
20260814 & --- & A: 12.37089; D: 12.34337 & $-0.02752$ \\
20260815 & --- & A: 12.18388; D: 12.02022 & $-0.16366$ \\
20260816 & --- & A: 12.46503; D: 12.32205 & $-0.14298$ \\
\bottomrule
\end{tabular}
\end{table}

All three paired seeds favor D on validation and sealed test. Test was evaluated
once after validation-selected checkpoints were frozen. Exact and
transposition-invariant families do not cross splits. Because all six final
curves still improve from epoch 4.5 to
epoch 5, the result establishes directional replication under a matched budget,
not convergence or a ComMU capability limit. On clean test, D primarily
improves Type and Time, leaves Duration nearly unchanged, and slightly worsens
Pitch. Because the A/D intervention holds pitch encoding fixed, this Pitch
difference is an indirect cross-factor effect rather than evidence for or
against pitch-coordinate construction. The replication therefore supports the
direction of the total temporal-interface effect, not universal improvement of
every factor head.

\end{document}